\documentclass[runningheads]{llncs}
\usepackage{graphicx}
\usepackage{hyperref}
\usepackage{graphicx}
\usepackage{multirow}
\usepackage{caption}
\usepackage{subcaption}
\begin{document}
\title{Towards real-time and energy efficient Siamese tracking -- a hardware-software approach}
\titlerunning{Towards real-time and energy efficient Siamese tracking}
%

\author{
Dominika Przewlocka-Rus\href{https://orcid.org/0000-0002-5836-8604}{\includegraphics[width=16pt]{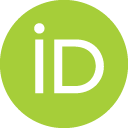}}
 \and Tomasz Kryjak\href{https://orcid.org/0000-0001-6798-4444}{\includegraphics[width=16pt]{images/orcid.png}}
}

%
\authorrunning{D. Przewlocka-Rus, T. Kryjak}

%
\institute{Embedded Vision Systems Group, Computer Vision Laboratory, \\ Department of Automatic Control and Robotics, \\ AGH University of Science and Technology, Krakow, Poland
\email{\{dominika.przewlocka,tomasz.kryjak\}@agh.edu.pl}}
\maketitle              
\begin{abstract}


Siamese trackers have been among the state-of-the-art solutions in each Visual Object Tracking (VOT) challenge over the past few years. 
However, with great accuracy comes great computational complexity: to achieve real-time processing, these trackers have to be massively parallelised and are usually run on high-end GPUs. 
Easy to implement, this approach is energy consuming, and thus cannot be used in many low-power applications. 
To overcome this, one can use energy-efficient embedded devices, such as heterogeneous platforms joining the ARM processor system with programmable logic (FPGA). 
In this work, we propose a hardware-software implementation of the well-known fully connected Siamese tracker (SiamFC). 
We have developed a quantised Siamese network for the FINN accelerator, using algorithm-accelerator co-design, and performed design space exploration to achieve the best efficiency-to-energy ratio (determined by FPS and used resources).
For our network, running in the programmable logic part of the Zynq UltraScale+ MPSoC ZCU104, we achieved the processing of almost 50 frames-per-second with tracker accuracy on par with its floating point counterpart, as well as the original SiamFC network.
The complete tracking system, implemented in ARM with the network accelerated on FPGA, achieves up to 17 fps.
These results bring us towards bridging the gap between the highly accurate but energy-demanding algorithms and energy-efficient solutions ready to be used in low-power, edge systems.


\end{abstract}
\section{Introduction}

Visual object tracking is a component of many different advanced computer vision systems, used, among others, in surveillance systems, advanced driver assistance systems (ADAS), or autonomous vehicles, such as cars or drones.
Due to the high complexity of the considered problem -- the tracked object can undergo changes, such as rotations, occlusions, or different scene illuminations -- it is still a research area of very high activity, well documented each year by the Visual Object Tracking Challenge (or VOT Challenge). 
Tracking methods can be roughly divided into classic (mean-shift, CAM-shift, KLT) and AI-based ones (including correlation filters). 
It is the development of deep learning that has allowed for significant progress in the field of tracking, and nowadays top trackers are based on neural networks, including the Siamese neural networks.
Unfortunately, state-of-the-art trackers are usually characterised with a very high computational complexity (resulting, inter alia, from the very fact of using neural networks) and to achieve real-time processing, these algorithms are accelerated using high-end and energy-inefficient GPUs.
At the same time, in many real-life applications, the real-time and energy-efficient processing constraints have to be met while ensuring high-quality tracking.
One of the possible solutions is the acceleration of state-of-the-art trackers using SoC FPGA (System on Chip Field Programmable Gate Arrays) platforms, which allow for high parallelisation of computations, with low energy consumption.
Nevertheless, this choice results in other challenges, mainly due to the limited number of resources in FPGA devices.
In view of the above, in this work we propose a fast and energy-efficient hardware-software implementation of the SiamFC tracker \cite{bertinetto2016fully}, achieving accuracy on par with that of its original counterpart.
The main contributions of this paper are:

\begin{itemize}
    \item The hardware-software implementation of a Siamese tracker on the Zynq UltraScale+ MPSoC ZCU104 platform, with a detailed time analysis of the algorithm components and a design space exploration showing the relation between the energy used (used resources) and the achieved speed (measured in FPS).
    \item The proposed algorithm-accelerator co-designed architecture of a Siamese neural network, which resulted in tracking accuracy on par with the original SiamFC approach, while significantly reducing the number of parameters (thus calculations).
    \end{itemize}
To the authors' best knowledge, this is the first paper to give such a comprehensive analysis of hardware-software implementation of the SiamFC tracker.

The remainder of this paper is organised as follows.
In Section \ref{sec:siam} we briefly describe the concept of Siamese trackers and discuss related work.
The proposed quantised Siamese tracker and its hardware-software implementation are presented in Section \ref{sec:quantized_tracker}.
The proposed tracker is evaluated both in software and hardware, with extensive accuracy, time and energy consumption analysis. The obtained results are discussed in Section \ref{sec:results}.
The paper ends with conclusion and future research proposals.

\section{Siamese Tracking}
\label{sec:siam}

A Siamese network is a Y-shaped network with two branches joined in one output.
It measures the similarity of the two inputs, thus it can be considered as a similarity function. 
Many of the Siamese-based trackers rely solely on this assumption.
In general, we examine the following two inputs: the exemplar image of an object (from the first frame) and the region of interest (ROI), where we presume that the target is present in the following frames.
Each branch processes one image, and their outputs are joined using correlation. 
This results in a similarity map (or maps) between object features and ROI, based on which the target can be located.
Over the past years, this basic version of the Siamese tracker has undergone many modifications, affecting both the tracking efficiency and frame processing time; the most recognisable are described in Section \ref{subsec:related_work}.


\subsection{Related Work} \label{subsec:related_work}

Fully convolutional Siamese trackers were first introduced in the paper \cite{bertinetto2016fully}. 
Both the object and ROI are processed by identical branches, based on the well-known AlexNet DCNN (Deep Convolutional Neural Network) architecture. 
The feature maps obtained are cross-correlated to produce a single heat map, determining the location of the target centre. 
The ROI is analysed in multiple scales -- the one for which the heat map has the highest peak is chosen to rescale the previous bounding box.
A direct continuation of the research is presented in \cite{Valmadre_2017_CVPR}, where the previous solution was extended with a correlation filter as an additional layer of the Siamese network.
To overcome the issue of too deep Siamese networks, in \cite{DensSiam}, a dense block-based network architecture was proposed. 
Each dense block is built of multiple convolution layers, the outputs of which are feed-forwarded to all next blocks. 
In this way, both low-level and high-level features are cross-correlated, which enhances the network's generalisation ability. 
In addition, the ROI branch was equipped with an attention module.

To avoid multiscale search, in \cite{SiamRPN} the Siamese-RPN framework was proposed. 
It consists of the Siamese network for features' extraction (ended with a~cross-correlation) and the two-branch Region Proposal Network: one for foreground-background classification and the other for proposal refinement.
In \cite{SiamRPN2} the extension of SiamRPN was proposed. 
The output of the Siamese network is extended to aggregate the outputs of the intermediate layers. 
This allows the similarity map to be calculated using features learnt on multiple levels. 
Moreover, the correlation layer is replaced with a~depth-wise separable correlation, which results in a multichannel similarity map with different semantic meanings for each channel. 
Also in this work, the authors proposed a different backbone than previously used (\cite{bertinetto2016fully}, \cite{Valmadre_2017_CVPR}, \cite{SiamRPN}) -- instead of AlexNet modification, they developed an appropriately adjusted ResNet50 and pointed out the conditions to be met by deep networks to be used in Siamese trackers.

In \cite{SiamVGG}, the authors proposed a VGG-16 like backbone and indicated that the most commonly used AlexNet has limited feature extraction capabilities.
Unlike \cite{SiamRPN} and \cite{SiamRPN2} based on anchors, as well as \cite{bertinetto2016fully}, \cite{Valmadre_2017_CVPR} with multiscale search, the solution proposed in \cite{SiamCAR} reformulated the tracking problem as a joint regression and classification task. 
Depth-wise correlation is applied to aggregated Siamese network output (with outputs from intermediate layers), and then the result is passed to two networks: one for foreground-background classification and the other for bounding box regression.
In \cite{SiamMask}, benefiting from depth-wise cross-correlation, the authors proposed a new approach with the network output representing the binary segmentation mask for the target. 
This also enabled the prediction of rotated bounding boxes.

The works listed above do not exhaust the progress that is constantly being made in the field of Siamese-based trackers, but serve more as an overview of the most important concepts: different backbones, dealing with scale changes with multiscale search, anchors or bounding box regression, and finally the analysis of output features of multiple levels. A comprehensive summary of existing algorithms with a description of current trends can be found in \cite{SiamTrackSurvey_2021}.

The listed algorithms are characterised with high tracking accuracy; however, to achieve real-time processing, they are run on high-end GPUs. 
There are only a few works on the low-power and real-time implementation of Siamese trackers.
In \cite{SiamICCVG} the authors presented preliminary results on the optimisation of fully connected Siamese networks for object tracking. 
With various experiments on quantisation and backbone architecture, they showed that precision reduction can positively affect the overfitting, thus also tracking accuracy. 
Similarly in \cite{Cao2019} the authors focused on optimisation of the size of the Siamese network's architecture which, however, significantly influenced the effectiveness of the tracker. 
On the other hand, in the paper \cite{Chen2020}, the results on effective co-design for algorithm and accelerator for AI on edge were presented, also for networks typically used in Siamese trackers.
In \cite{SiamPYNQ} the authors proposed a hardware-software implementation of a SiamRPN-like tracker in PYNQ (ZCU 104). 
The PS (Processor System) part is used for system configuration, reading the input frames, communication with accelerator, and displaying the results. 
In PL (Programmable Logic) two networks are accelerated: Siamese and Region Proposal. 
The authors report that their tracker runs with $36.7$ FPS. 
Unfortunately, in the cited paper, there is no information on resource usage or energy consumption, as well as it is not clearly stated what network architecture was used and if (and how) it was quantised, which both have a direct impact on hardware implementation feasibility. 
Moreover, the tracker accuracy is not provided, nor is the comparison with the baseline (software) solution.
Similarly in \cite{Zhang2018} the authors proposed the hardware implementation of a lightweight Siamese network using both pruning and quantisation. The tracker, running on ZedBoard with a ZCU 102 core, achieves $18.6$ FPS. 
However, since the article lacks a description of other than network accelerator components, that is, acquisition of input data or post-processing of network's output to obtain the location of the target, it seems that the FPS rate refers solely to the neural network (not the complete tracking system).

On the basis of this analysis, one can notice a significant progress in the Siamese tracker domain (measured mainly by trackers' accuracy), which, however, is not accompanied by equally extensive research on their embedded devices' deployment. The few existing works on hardware acceleration of Siamese trackers lack of important details which makes them hard to compare.
At the same time, since in many applications we face the challenge of real-time and energy-efficient processing, choosing such hardware may be necessary. 
This directly motivates our research.

.



\section{Quantised Siamese Tracker} 
\label{sec:quantized_tracker}


The most commonly used backbones for Siamese trackers are appropriate modifications of AlexNet or ResNet networks. 
Nevertheless in case of embedded devices implementations, one of the key elements in network architecture selection is most of all the accelerator design and its limitations (resulting also from the limited on-board resources). 
In this work, for network acceleration, we use the FINN framework (\cite{blott2018finn}, \cite{finn}) and the Zynq UltraScale+ MPSoC ZCU104 platform. 
For algorithm-accelerator co-design, in particular, one has to take into account the following factors:
\begin{itemize}
    \item Computations precision, which results directly from the number of bits for the coding of the weights and activations. Apart from the reduction of needed memory, this also affects the number of resources used for arithmetical operations (e.g. floating point operations are far more complex than 8- or even 4-bit integer ones).
    \item Unified and small filters positively affect the possibility of computations parallelisation.
    \item Using a too deep network architecture can negatively affect the possibility of parallelisation (or, in the extreme case, cannot be implemented on a given platform).
    \item Custom and specific architectures may not be supported for the chosen FINN accelerator.
    \item Careful tuning of the folding parameters can increase parallelisation, thus decrease processing time, but at the cost of the number of resources used.
\end{itemize}

Given the above, we have designed a custom Siamese network architecture presented in Table \ref{tab:net_arch} (one branch) -- for all layers, we used zero padding and stride equal to $1$. 
After each convolution layer, except for the last, there is a batch normalisation layer. 
The input image, the ROI, is of size 238x238x3, while the one representing the object to track (used for initialisation) is of size 110x110x3. 
The activations are quantised to 4 bits, while using 8 bits precision for weights of the first and last layers allows to maintain high accuracy.
\begin{table}[!t]
\begin{center}
\caption{The proposed network architecture}\label{tab:net_arch}
\setlength{\tabcolsep}{6pt}
\begin{tabular}{ l c c c c }
\hline
Layer & Kernel & Filters No. & Quantisation & Maxpooling \\
\hline
Conv 1.1 & 3x3 & 64 & 8 bits & 2x2 \\
Conv 1.2 & 3x3 & 64 & 4 bits & 2x2 \\
Conv 2 & 3x3 & 128 & 4 bits & 2x2 \\
Conv 3 & 3x3 & 128 & 4 bits & - \\
Conv 4 & 3x3 & 128 & 4 bits & - \\
Conv 5 & 3x3 & 128 & 8 bits & - \\
\hline
\end{tabular}
\end{center}
\end{table}

The proposed tracking algorithm is based on SiamFC \cite{bertinetto2016fully}, which does not use the aggregation of outputs from intermediate layers, additional branches for classification or bounding box regression and other complex elements (for full algorithm description please refer to the original work). 
It is especially important given the choice of a FINN accelerator, which does not support most of these operations straightforwardly. 
Obviously, this constrains the possibility of accelerating the best existing tracker and will be widely commented on in Sections \ref{sec:results} and \ref{sec:conclusion}.
Still, FINN allows to adapt the accelerator architecture to the chosen network (unlike e.g. Vitis AI) -- based on the properly prepared network graph and folding parameters, the hardware (accelerator) is generated. 
Folding parameters control the level of computations parallelisation: for each layer, we can set the number of simultaneously processed input channels (PE parameter) and the number of aggregated output channels (SIMD).

\subsection{Hardware-software implementation}

\begin{figure}[!t]
    \centering
    \includegraphics[width=1\textwidth]{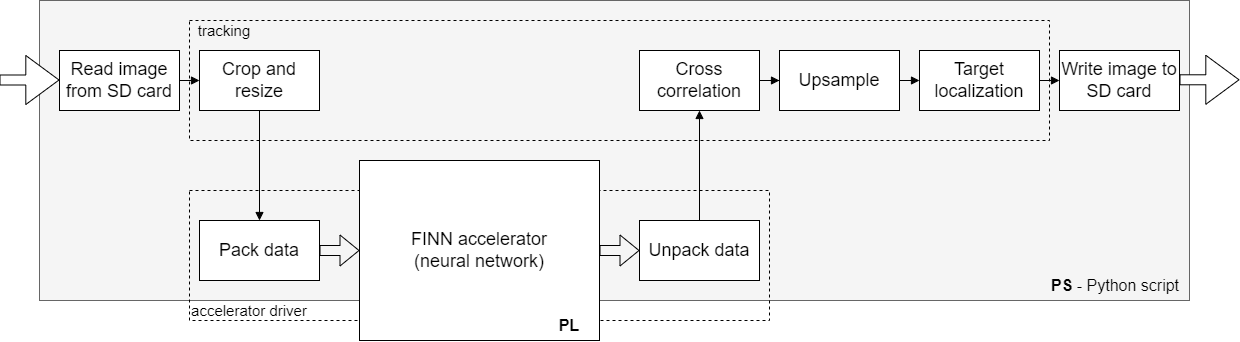}
    \caption{Overview of the proposed hardware-software system. A single branch of the Siamese network is accelerated using the FINN framework in PL (FPGA). The Python script is run on the ARM processor (PS), handling the input and output, communicating with the accelerator and post-processing the network output.}  
    \label{fig:system_implementation}
\end{figure}

In this paper, we use the hardware-software approach and divide the implementation of the tracker into the network accelerated in PL and the rest of the tracking algorithm implemented in PS. The Python script is run in ARM, which is responsible for: (1) input and output handling; (2) communication with FPGA via a~proper driver; (3) realisation of the tracker's logic -- cropping and scaling the input image, and then post-processing the output of the network (determining the target location based on similarity map). An overview of the proposed system is presented in Fig. \ref{fig:system_implementation}.
The software part of our tracker (in an ARM processor) is run with Python 3 interpreter, using numpy, PyTorch, and OpenCV libraries. The clock for the PL is set to 100 MHz.

\section{Results} 
\label{sec:results}

We have evaluated the proposed solution in two ways. 
Firstly, we have tested the developed quantised tracker on different datasets and compared its accuracy with the baseline model, as well as the original solution. 
Second, we have done a~design space exploration with different hardware settings to analyse the tracker performance. 
The details of these experiments are summarised below.

\subsection{Benchmark results}


To properly evaluate the proposed tracker, we have prepared two versions of the network described in Section \ref{sec:quantized_tracker}: floating point baseline and quantised. 
Both networks were trained in the GOT 10k dataset \cite{Huang2021} (unlike the original SiamFC tracker, trained on ImageNet), for 50 epochs, with an initial learning rate 1e-2, reduced each epoch to a final value of 1e-5.
Next, we conducted multiple experiments to compare the tracker accuracy for different scenarios: floating-point network, quantised network, and processing of single or three scales. The tracker was evaluated on the VOT 2016 dataset for a proper comparison with the original SiamFC \cite{bertinetto2016fully}.

%
\begin{table} [!t]
\begin{center}
\caption{Comparison of the performance of the tracker. The results were obtained using the GOT 10k toolkit. The mean average overlap (mAO) metric takes into account the potential class imbalance in the evaluation by updating the standard AO (denoting the average overlaps between all ground-truth and estimated bounding boxes) with weights proportional to the number of frames in each sequence \cite{Huang2021} (s -- scale)}
\label{tab:tracker_accuracy}
\setlength{\tabcolsep}{3pt}
\begin{tabular}{ l c }
\hline
\multirow{2}{*}{Tracker} & VOT 2016\\
& mAO \\
\hline
FP32 3s & $0.362$ \\
FP32 1s & $0.315$ \\
Quantised 3s & $0.355$\\
Quantised 1s & $0.281$ \\
Original SiamFC \cite{bertinetto2016fully} (3s) & $0.385$ $\dagger$ \\
\hline
\multicolumn{2}{l}{\tiny{$\dagger$ raw results downloaded from official VOT2016 challenge \cite{Kristan2016a}}} \\
\end{tabular}
\end{center}
\end{table}
Table \ref{tab:tracker_accuracy} summarises the obtained results (we do not present the comparison with other Siamese tracker FPGA accelerators since in previous works -- summarised in Sec. \ref{subsec:related_work} -- authors either do not report any accuracy results, or use other metrics):
\begin{itemize}
    \item For the VOT 2016 benchmark with the proposed network, our tracker achieves accuracy on par with the original SiamFC \cite{bertinetto2016fully} when processing 3 scales regardless of the quantisation: the FP32 3s tracker is around $6\%$ behind the original, while the quantised 3s around $8\%$. However, it is crucial to notice that the proposed network is far more compact than the AlexNet-based one. Specifically, the AlexNet backbone has $3747200$ parameters, while ours $554688$, which is around $6.7x$ less.
    \item For our tracker, the best accuracy (measured by mAO) is obtained using the FP32 3 scale network. Nevertheless, after quantisation we observe only a slight decrease in accuracy - from $0.362$ to $0.355$ (less than $2\%$).
    \item The decrease in accuracy is greater after reducing the number of processed scales from 3 to 1. For the FP32 network, the difference is around $0.032$ ($9\%$), while for the quantised one, even $0.074$ (around $21\%$).
\end{itemize}
\begin{figure}[!t]
\begin{center}
\captionsetup[subfigure]{labelformat=empty}
\subfloat[Initialization]{\includegraphics[width=0.25\textwidth]{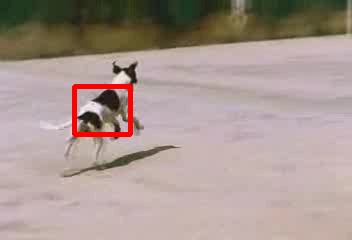}} \hfill
\subfloat[Frame 10]{\includegraphics[width=0.25\textwidth]{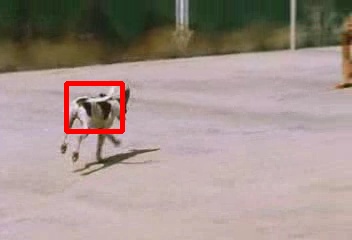}} 
\subfloat[Frame 50]{\includegraphics[width=0.25\textwidth]{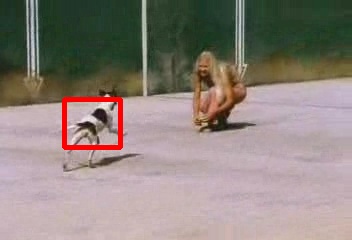}} \hfill
\subfloat[Frame 100]{\includegraphics[width=0.25\textwidth]{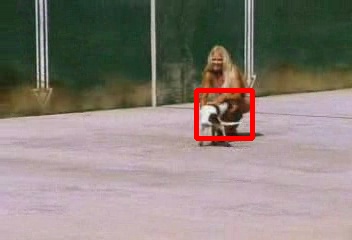}} 
\caption{Output from the quantised Siamese 1 scale tracker for 'Dog' sequence from OTB.}
\label{fig:exemplar_output}
\end{center}
\end{figure}
Figure \ref{fig:exemplar_output} shows an exemplar output of the quantised 1 scale tracker.

\subsection{Performance}
To obtain the best network acceleration performance using FINN we performed a design space exploration for choosing the right folding parameters (see Sec. \ref{sec:quantized_tracker}). 
The set parameters for six different experiments are summarised in the Table \ref{tab:folding}. 
\begin{table}[!t]
\begin{center}
\caption{Different folding settings for the FINN accelerator}\label{tab:folding}
\setlength{\tabcolsep}{6pt}
\begin{tabular}{ l c c c c c c }
\hline
 & \multicolumn{6}{c}{Layers' (PE, SIMD)} \\
 & 1 & 2 & 3 & 4 & 5 & 6\\
\hline
V1 & (32, 3) & (32, 16) & (16, 16) & (8, 16) & (8, 16) & (8, 8) \\
V2 & (32, 3) & (32, 16) & (16, 16) & (8, 16) & (8, 16) & (16, 8) \\
V3 & (32, 3) & (32, 16) & (16, 16) & (16, 16) & (16, 16) & (16, 8) \\
V4 & (32, 3) & (32, 16) & (16, 16) & (16, 16) & (16, 16) & (16, 16) \\
V5 & (32, 3) & (32, 16) & (32, 16) & (32, 16) & (32, 16) & (32, 16) \\
V6 & (32, 3) & (32, 16) & (32, 16) & (32, 32) & (32, 32) & (32, 32) \\
\hline
\end{tabular}
\end{center}
\end{table}
After hardware generation we analysed the used resources, energy consumption, and latency of the network input processing. The results are summarised in the Table  \ref{tab:accelerator_performance}.
\begin{table}[!t]
\begin{center}
\caption{Comparison of accelerated Siamese network performance for different folding configurations. When increasing the level of parallelisation (using the number of PEs and SIMDs), we can observe both an increase in processing speed and energy consumption}\label{tab:accelerator_performance}
\setlength{\tabcolsep}{6pt}
\begin{tabular}{ l c c c c c c }
\hline
\multirow{2}{*}{Folding} & \multicolumn{4}{c}{Resources} & \multirow{2}{*}{FPS} & \multirow{2}{*}{Energy [W]} \\
 & LUT & FF & BRAM & LUTRAM &  & \\
\hline
V1 & 40.45\% & 16.78\% & 46.31\% & 11.92\% & $38.46$ & $4.5$\\
V2 & 42.25\% & 17.44\% & 50.8\% & 12.1\% & $40.24$ & $4.56$\\
V3 & 46.66\% & 17.9\% & 50.8\% & 12.1\% & $41.31$ & $4.81$\\
V4 & 48.72\% & 18.6\% & 50.8\% & 12.16\% & $42.16$ & $4.92$\\
V5 & 66.87\% & 23\% & 55.29\% & 12.54\% & $49.03$ & $5.5$\\
V6 & 91.27\% & 28.66\% & 91.83\% & 13.58\% & $49.63$ & $6.79$\\
\hline
\end{tabular}
\end{center}
\end{table}
%

\begin{figure}[!t]
    \centering
    \includegraphics[width=0.55\textwidth]{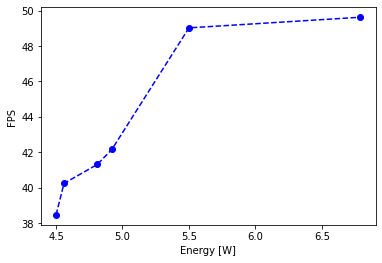}
    \caption{Design space exploration for acceleration of Siamese network with energy to FPS ratio. The used energy is tightly connected to the used resources presented in Table \ref{tab:accelerator_performance}.}
    \label{fig:design_space_exploration}  
\end{figure}

For experiments V1, V2, V3 and V4 we gradually increase the number of, first, PE elements, and then SIMD, which results in a slight decrease in the used resources: mainly LUTs, responsible for arithmetical operations, but also BRAMs. 
At the same time, we observe a stable but subtle increase in FPS, from $~38$ to $~42$, with a simultaneous increase in energy consumption of around $0.5$ W.
A considerable change was achieved after doubling the number of PEs in layers 3, 4, 5, 6 -- for experiment V5, in relation to V4. 
The number of LUTs used increased by around $18\%$, BRAMs $5\%$, FFs $5\%$, which accelerated processing by around $7$ FPS to $49$ FPS, with an increase in energy consumption of $~0.6$ W.
Interestingly, next experiments with increasing the level of parallelisation -- V6, where we double the number of SIMDs for layers 4, 5, 6 -- caused considerable increase in the used resources (over $90\%$ available LUTs and BRAMs) and the energy consumption to almost $7$W, while improving processing speed by only $0.6$ FPS.
The dependence between energy consumption and the FPS achieved is presented in Figure \ref{fig:design_space_exploration}.
%

Table \ref{tab:accelerator_performance} shows the results only for network acceleration and does not take into account transfers to and from accelerator, as well as the rest of the tracker computations. 
Careful time analysis for the complete software-hardware implementation, using the V5 accelerator, is presented in Tab. \ref{tab:tracker_time}, from which we draw several conclusions:
\begin{itemize}
    \item The network acceleration takes $50\%$ of the time needed to process the ROI. This also includes the time for packing the input data, transferring them to the accelerator, from accelerator, and unpacking. In other words, acceleration with data transfer enables the processing of a single ROI with a speed of around $35$ FPS.
    \item Almost half the time for post-processing of the network output (around $14\%$ of total) is used for cross-correlation, while around $10\%$ for target location (with, among others, cosine window filtration).
    \item The network input pre-processing (cropping and scaling ROI) takes relatively little time. Much greater impact on the processing speed is the transfer of data from the accelerator (and unpacking) than the transfer to the accelerator (including packing) -- $14\%$ vs $2\%$ of total time. 
\end{itemize}
Finally, the complete hardware-software tracking system processes a frame with around $17$ FPS.
%
%
\begin{table}
\begin{center}
\caption{Analysis of the average latency of each tracking stage for the V5 folding version. The network acceleration with I/O data transfers achieves around $35$ FPS, while the complete tracking system operates at a speed of $~17$ FPS.}\label{tab:tracker_time}
\setlength{\tabcolsep}{6pt}
\begin{tabular}{ l c}
\hline
Stage &  Time [ms]\\
\hline
Crop \& resize & $0.0102$ \\
Input transfer & $0.001$ \\
Network acceleration & $0.0205$ \\
Output transfer & $0.008$ \\
Cross correlation & $0.0081$ \\
Upsampling & $0.0011$ \\
Locating target & $0.0057$ \\
\textbf{Sum} & $\mathbf{0.0546}$ \\
Total (measured)* & $0.0587$ \\
\hline
Input preprocessing & $0.0102$ (18\%)\\
FINN network transfer \& execution & $0.0295$ (52\%) \\
Network output processing & $0.0149$ (25\%) \\
\hline
\multicolumn{2}{l}{\tiny{*with other additional operations}}
\end{tabular}
\end{center}
\end{table}

\section{Conclusion} \label{sec:conclusion}


In this work, we have proposed a hardware-software implementation of a Siamese tracker, based on \cite{bertinetto2016fully}. Firstly, we have designed a Siamese neural network, which architecture meets the chosen FINN accelerator constraints, and at the same time allows our tracker to achieve the accuracy on par with the original SiamFC solution, even for the quantised version. 
Second, we have performed a design space exploration, increasing the level of paralellisation in FINN accelerator and have shown the relation between the energy consumption and tracker speed. Finally, our tracker achieves around $17$ FPS with $5.5$ W energy consumption. The original tracker run on NVIDIA GeForce GTX Titan X with $250$ W energy consumption, achieved $83$ FPS \cite{bertinetto2016fully}.
We have also provided a time analysis of each tracker component and pointed out the bottlenecks of the proposed solution.
On the basis of that, we draw two main conclusions for future work:
\begin{itemize}
    \item Despite the fact that our tracker is on par with the original SiamFC, the accuracy achieved is far behind the best existing Siamese tracking algorithms. The next work should then be supplemented with different SoTA features, such as bounding box regression or aggregation of features from different levels. It is important to note that acceleration of such a network would not be possible using FINN in a straightforward manner. Therefore, for future work, it is planned to usethe FINN accelerator as one of the hardware components, along with some other, custom solution for e.g. bounding box regression.
    \item In the current version of the tracker, data transfer from the accelerator and output analysis have a big impact on the latency of the solution (almost $40\%$ of total time). Moving the post-processing to FPGA would significantly improve the frame processing time, since the output transfer to PS would not be needed, and, at the same time, the cross correlation could be parallelised.
\end{itemize}
%
Based on the above, we also want to pay attention to the fact that the progress in developing more and more accurate tracking algorithms (including the Siamese-based ones) is far beyond the progress in AI on edge deployment, especially for available, ready-to-use accelerators. Such solutions usually do not support various advanced methods standard for software approaches, at least not without a deep interference in the source code (which is still feasible only for rare open-source solutions). 
Faced with the need to significantly reduce the energy demand, both for deployment on low-power devices and for global needs, we believe that continuous work on energy-efficient advanced vision systems is especially important.

\section*{Acknowledgements}
The work presented in this paper was supported by the National Science Centre project no. 2016/23/D/ST6/01389 entitled ``The development of computing resources organisation in latest generation of heterogeneous reconfigurable devices enabling real-time processing of UHD/4K video stream. The authors would like to thank Joanna Stanisz and Konrad Lis for they supprort when working with FINN, in particular on the reduction of the data transfer time to and from the accelerator 

%
%
\bibliographystyle{splncs04}
%

\end{document}